\documentclass[letterpaper, 10 pt, conference]{ieeeconf}
\IEEEoverridecommandlockouts
\usepackage{amsmath,amsfonts}
\usepackage{subfigure}
\usepackage{graphicx}
\usepackage{balance}
\usepackage{algorithm}
\usepackage{algpseudocode}
\usepackage{booktabs}
\usepackage{threeparttable}
\usepackage{hyperref}
\usepackage{array}
\usepackage{textcomp}
\usepackage{stfloats}
\usepackage{url}
\usepackage{verbatim}
\usepackage{cite}
\usepackage{algpseudocode}
\usepackage{stfloats}
\usepackage{microtype}
\usepackage{multirow}

\usepackage{amsmath,amssymb}
\usepackage{algorithm}
\usepackage{algpseudocode}

\allowdisplaybreaks

\begin{document}

\setlength{\abovedisplayskip}{4pt}
\setlength{\belowdisplayskip}{4pt}

\title{\bfseries \parbox{0.85\linewidth}{\centering Experience-Guided Initiation Search for Learned Skills in Skill Composition }}

\author{ Qixuan Li$^{1}$, Yanhong Zhao$^{2}$, Jincheng Yu$^{3}$
\thanks{$^1$ Shenzhen International Graduate School, Tsinghua University, Shenzhen, China. lqx23@mails.tsinghua.edu.cn.} 
\thanks{$^2$ School of Electronic Science and Engineering, Nanjing University, Nanjing, China. 231870178@smail.nju.edu.cn.}
\thanks{$^3$ Department of Electronic Engineering, and the Institute for Embodied Intelligence and Robotics, Tsinghua University, Beijing, China. yu-jc@mail.tsinghua.edu.cn. }
}



\maketitle

\begin{abstract}
Deploying frozen learned skills, such as Vision-Language-Action (VLA) policies, in new environments requires identifying initiation configurations that support reliable execution. In skill composition, an initiation configuration affects not only the current skill but also the physical state passed to subsequent skills, so successful execution of an individual skill does not necessarily imply successful completion of the composed task. Estimating target-specific capability through extensive rollouts is costly in real-world deployment, while directly reusing historical experience can be unreliable under environment changes. We propose EVIS, an Experience-Guided and Behavior-Validated Initiation Search framework for discovering reliable initiation configurations under limited target interaction. EVIS uses historical execution experience to prioritize promising candidates and target-environment behavior to validate whether they remain effective. We evaluate EVIS on single-skill and two-stage manipulation tasks with frozen VLA policies. EVIS reduces mean target-environment queries and improves reliable candidate discovery under small interaction budgets. These results show that combining historical guidance with target-side behavioral validation can reduce the interaction cost of deploying frozen learned skills in new environments.
\end{abstract}


\section{Introduction}
\label{sec:Introduction}

Learned visuomotor skills, including pretrained Vision-Language-Action (VLA) policies, provide reusable manipulation capabilities and are increasingly composed in hierarchical systems for longer-horizon tasks~\cite{kim2024openvla,huang2026roboharness,zhang2026harness,huang2022inner}.
However, a symbolically valid skill sequence is not necessarily physically executable, because the initiation configuration of an upstream skill affects both its own execution and the physical state handed to subsequent skills.
An upstream skill may therefore succeed locally yet leave the system in a state from which the next skill cannot proceed; for example, successfully opening a drawer does not guarantee successful subsequent retrieval or placement.
Thus, composing frozen learned skills requires determining not only \emph{which skill to invoke}, but also \emph{from which initiation configuration it should be executed}.

Determining such initiation configurations in a new environment is challenging.
Existing Task and Motion Planning (TAMP) methods primarily describe action applicability through geometric constraints and symbolic preconditions~\cite{garrett2021integrated,garrett2020pddlstream}, while learned precondition models estimate skill applicability from execution data~\cite{liang2022learning,sharma2020relational}.
Recent memory-based methods further reuse historical execution experience to assist the invocation and composition of frozen policies~\cite{huang2026roboharness,zhang2026harness}.
However, for a frozen black-box skill in a new environment, the reliability of different initiation configurations is generally unknown.
Estimating target-specific capability through extensive rollouts requires substantial interaction, while directly reusing historical candidates may fail as environment geometry, visual appearance, and interaction conditions change.
Historical experience is therefore useful as guidance, but insufficient as a decision criterion.
This leads to the central question of this work:
\emph{how can historical experience help discover reliable initiation configurations with fewer target interactions, while remaining robust to imperfect transfer?}

\begin{figure}[t]
    \centering
    \includegraphics[width=\columnwidth]{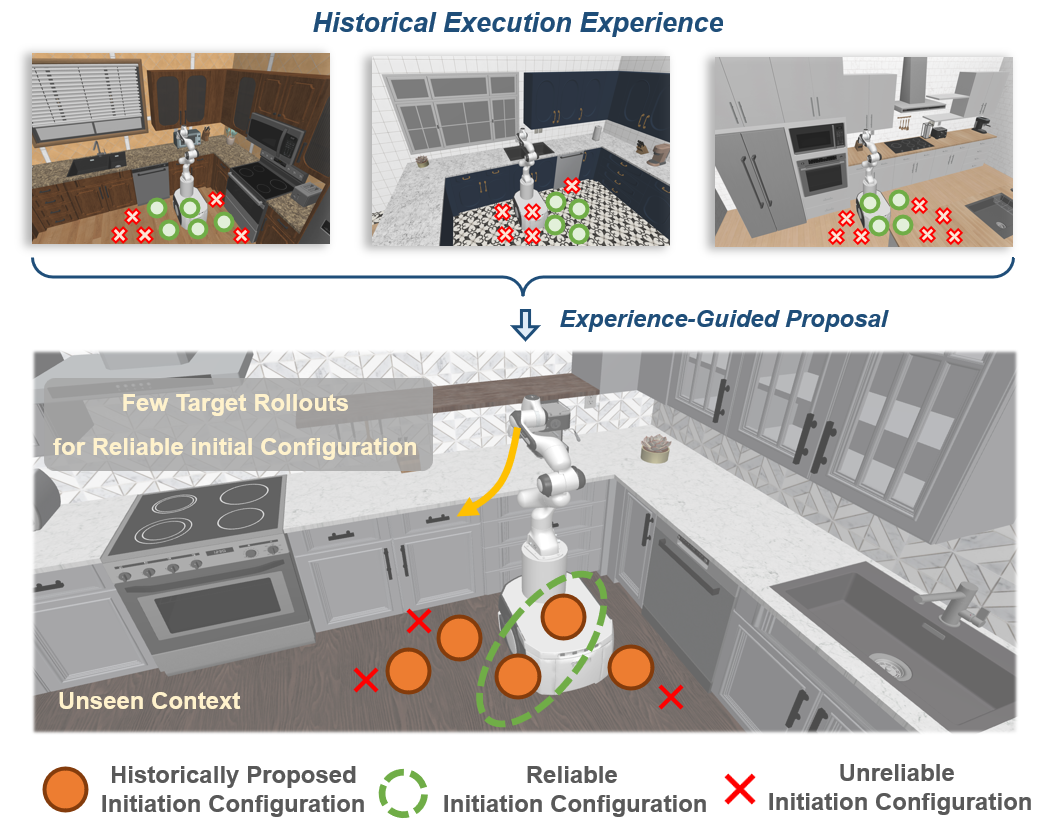}
    \caption{
        \textbf{Experience-guided and behavior-validated initiation search for frozen learned skills.}
        Historical execution experience from prior contexts is used to propose promising initiation configurations for an unseen target context.
        EVIS then uses a small number of target-environment rollouts to validate these proposals and identify a reliable initiation configuration for the desired execution objective.
    }
    \label{fig:overview}
    \vspace{-0.3cm}
\end{figure}

To address this problem, we propose \textbf{EVIS}, an \textbf{E}xperience-Guided and \textbf{B}ehavior-Validated \textbf{I}nitiation \textbf{S}earch framework.
EVIS treats historical experience as a source of candidate proposals rather than final decisions.
Historical execution records prioritize promising initiation configurations, while behavioral evidence from target-environment rollouts---i.e., observed outcomes under the intended execution objective---validates whether these candidates remain effective.
This separates transfer from decision-making: history determines where to look first, whereas target-side outcomes determine whether a candidate should be trusted.
When sufficient behavioral evidence supports a candidate, the search can terminate early; otherwise, it continues to seek alternatives.

We evaluate EVIS in RoboCasa365~\cite{nasiriany2026robocasa365} using a frozen pretrained VLA policy~\cite{rldx2026}, on both single-skill and two-stage manipulation tasks.
Two-stage composition provides the minimal setting that isolates inter-skill handoff effects: a single skill contains no handoff, whereas adding one downstream skill is sufficient to test whether an upstream initiation configuration leaves a physical state that supports subsequent execution.
Across unseen environments, historical guidance improves early candidate discovery under small interaction budgets.
Behavioral validation filters unreliable transferred candidates, while continued search recovers part of the coverage lost when transfer is rejected.

Our contributions can be summarized as:
\begin{enumerate}

    \item We treat the initiation configuration as an explicit execution interface for task-level planning over frozen learned skills, and seek reliable configurations in new environments with as few costly target-environment rollouts as possible.

    \item We propose EVIS, an Experience-Guided and Behavior-Validated Initiation Search framework that uses historical execution experience to prioritize promising candidates, while target-environment rollouts determine whether a candidate should be accepted or the search should continue.

    \item We evaluate EVIS on RoboCasa365 environments using a frozen pretrained VLA policy. EVIS reduces target-environment interaction and improves reliable candidate discovery under small interaction budgets.
\end{enumerate}
\section{Related Work}
\label{sec:related_work}

\subsection{Skill Applicability and Composition}

Skill applicability and composition have long been studied in hierarchical control and planning.
The options framework formalizes skill initiation and termination~\cite{sutton1999between}, while skill chaining learns skills that reach the initiation sets of subsequent skills~\cite{konidaris2009skill}.
In task and motion planning, PDDLStream integrates symbolic planning with black-box samplers for continuous feasibility constraints~\cite{garrett2020pddlstream}.
Learned precondition models further estimate manipulation applicability from execution data using object relations~\cite{sharma2020relational} or point-cloud representations for contact-rich controllers~\cite{liang2022learning}.
Unlike these approaches, which primarily model skill applicability or feasibility, EVIS searches for effective initiation configurations of frozen learned skills under limited target interaction, evaluating them by whether they support successful execution of the composed skill sequence.

\subsection{Policy Orchestration and Execution Memory}

Language-guided robotic systems connect high-level task reasoning with low-level execution in different ways.
SayCan grounds skill selection in learned affordances~\cite{ahn2022can}, Inner Monologue incorporates online execution feedback into high-level planning~\cite{huang2022inner}, and Code as Policies composes perception and control interfaces through generated programs~\cite{liang2023code}.
More recent execution-memory systems reuse past experience to organize and invoke frozen VLA primitives.
HarnessVLA reuses task-specific execution traces and persistent execution memory to guide primitive composition across deployment variations~\cite{zhang2026harness}, while RoboHarness uses execution memory and online evidence for capability-aware policy routing and compatible skill handoffs~\cite{huang2026roboharness}.
EVIS addresses a complementary problem: it transfers execution experience across environments specifically to guide the search over initiation configurations, while using a small number of target-environment rollouts to validate whether transferred candidates remain reliable.

\subsection{Experience-Guided Search and Candidate Discovery}

Historical information can accelerate expensive black-box search.
Intelligent Trial and Error uses a precomputed behavior repertoire for rapid robot adaptation~\cite{cully2015robots}, Most Likely Expected Improvement combines prior likelihood with expected improvement for data-efficient direct policy search~\cite{pautrat2018bayesian}, and ranking-weighted Gaussian process ensembles transfer previous optimization runs by weighting historical models according to their agreement with target observations~\cite{feurer2018scalable}.
Relatedly, threshold-based pure exploration in multi-armed bandits studies how to identify satisfactory candidates with limited samples: the thresholding bandit problem identifies arms whose expected rewards exceed a prescribed threshold under a fixed budget~\cite{locatelli2016optimal}, while good-arm identification (GAI) seeks such arms using as few samples as possible~\cite{kano2019good}, with recent work extending GAI to anytime and fixed-budget settings~\cite{jourdan2026anytime}.
EVIS builds on these ideas for initiation search of frozen learned skills, using execution experience from previous environments to prioritize candidates and limited target-environment rollouts to validate whether transferred candidates remain reliable.
\begin{figure*}[t]
    \centering
    \includegraphics[width=\linewidth]{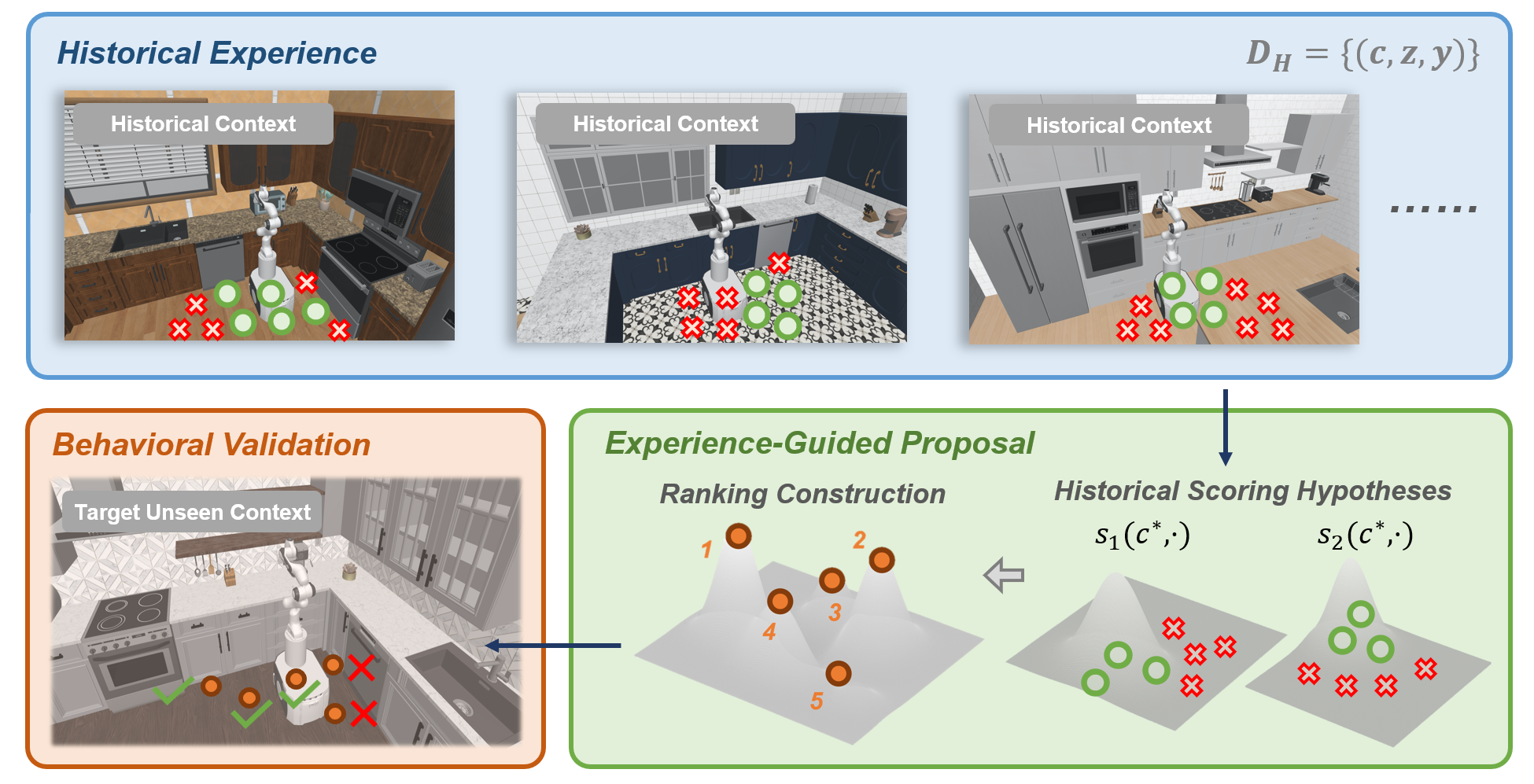}
    \caption{
    Overview of EVIS.
    Historical execution experience from prior contexts is used to
    construct multiple scoring hypotheses over initiation
    configurations, which provide complementary proposals for a new
    unseen target context.
    Target-environment rollouts then behaviorally validate the
    proposed configurations under objective $G$ and determine whether
    a candidate should be accepted or the search should continue.
    The underlying learned skill policy remains frozen throughout the
    process.
}
    \label{fig:framework}
\end{figure*}

\section{Problem Formulation}
\label{sec:problem_formulation}

We study initiation-configuration search for frozen learned skills in both single-skill and multi-stage manipulation.
Let $c \in \mathcal{C}$ denote an environment context and
$z \in \mathcal{Z} \subset \mathbb{R}^{d}$ denote the controllable
parameters of an initiation configuration.
The context $c$ describes the environment conditions, while $z$
specifies the robot--environment configuration from which execution
begins; for example, in mobile manipulation, $z$ may parameterize the
robot base pose relative to the target object or fixture.
For each problem instance, the underlying policies are fixed.

Let $G(\xi)\in\{0,1\}$ denote a binary success predicate over an
execution trajectory $\xi$.
For a single skill, $G(\xi)=1$ denotes successful skill completion;
for skill composition, it denotes successful completion of the full
sequence from the physical states produced by preceding skills.
We define the corresponding success probability as
\begin{equation}
    p_G(c,z)
    =
    \Pr\!\left(G(\xi)=1 \mid c,z\right).
    \label{eq:behavioral_capability}
\end{equation}
In a new environment, $p_G(c,z)$ is unknown, and each rollout provides
only a stochastic observation
\begin{equation}
    Y_G = G(\xi)
    \sim
    \operatorname{Bernoulli}\!\left(p_G(c,z)\right).
    \label{eq:behavioral_observation}
\end{equation}

We assume access to historical execution experience from $N$
previously encountered contexts:
\begin{equation}
    \mathcal{D}_{H}
    =
    \{(c_i,\mathcal{D}_i)\}_{i=1}^{N},
    \qquad
    \mathcal{D}_i
    =
    \{(z_{ij},y_{ij})\}_{j=1}^{n_i},
    \label{eq:historical_data}
\end{equation}
where $z_{ij}$ is an evaluated initiation configuration and
$y_{ij}\in\{0,1\}$ is its observed outcome under objective $G$.
These records characterize execution in previous contexts but need not
transfer reliably to a new target context $c^\star$.

After $t-1$ target-environment interactions, let
\begin{equation}
    \mathcal{D}_{t-1}
    =
    \{(z_k,y_k)\}_{k=1}^{t-1}
\end{equation}
denote the accumulated target-side observations.
The next initiation configuration is selected according to
\begin{equation}
    z_t
    =
    \mathcal{A}_t
    \left(
        \mathcal{D}_H,
        \mathcal{D}_{t-1}
    \right),
    \label{eq:adaptive_query}
\end{equation}
where $\mathcal{A}_t$ is the search policy.
Under a budget of at most $K_{\max}$ target executions, the procedure
returns either an initiation configuration $\hat z\in\mathcal{Z}$ or
abstains with $\hat z=\bot$.

For evaluation, given a reliability threshold $\rho\in(0,1)$, define
the latent reliable initiation set
\begin{equation}
    \mathcal{Z}_{\mathrm{rel}}(c^\star)
    =
    \left\{
        z \in \mathcal{Z}
        \mid
        p_G(c^\star,z) \geq \rho
    \right\}.
    \label{eq:reliable_initiation_set}
\end{equation}
Our goal is not to reconstruct the full target-specific success
function $p_G(c^\star,z)$, but to use historical experience and limited
target-side feedback to return a configuration in
$\mathcal{Z}_{\mathrm{rel}}(c^\star)$ with as few target executions as
possible.

From a pure-exploration perspective, this problem is related to
good-arm identification (GAI)~\cite{kano2019good}.
Under a finite candidate set, each initiation configuration can be
viewed as an arm with mean
$\mu_z=p_G(c^\star,z)$, with $\mu_z\geq\rho$ defining a good arm.
Unlike standard GAI, the search also has access to historical execution
data from related contexts, which can guide early queries while
target-side observations validate whether transferred candidates remain
reliable.
EVIS can therefore be viewed as a transfer-aware, budgeted good-arm
identification procedure over initiation configurations.
\section{Methods}
\label{sec:methods}

We propose \textbf{EVIS}, an Experience-Guided and Behavior-Validated
Initiation Search framework for the budgeted initiation-search problem
defined in Sec.~\ref{sec:problem_formulation}.
EVIS separates the roles of historical experience and target-side
interaction through three complementary mechanisms.
\emph{Experience-guided proposal} uses previous executions to prioritize
where to search first;
\emph{behavioral validation} determines whether target-side evidence is
sufficient to return a provisional candidate;
and \emph{behavior-guided recovery} continues the search when transferred
proposals are unsuccessful or insufficiently supported.
Thus, historical experience guides the ordering of candidate evaluations,
whereas target-environment behavior determines whether a candidate should
be trusted or whether additional search is required.

\subsection{Experience-Guided Proposal}
\label{subsec:historical_proposals}

EVIS constructs historical proposals over a finite candidate set
$\mathcal{Z}_{\mathrm{cand}}\subset\mathcal{Z}$.
To make historical experience comparable across environments,
we parameterize initiation configurations relative to the target fixture
and normalize each dimension by its search range.
Historical predictors are then evaluated on
$\mathcal{Z}_{\mathrm{cand}}$ to rank candidate configurations,
even when the original historical observations were collected at
different configurations.

\textbf{Historical scoring hypotheses.}
Historical experience may provide several useful but imperfect views of
which initiation configurations are promising in a new context.
We represent these views as a collection of historical scoring
hypotheses
\begin{equation}
    \mathcal{H}_H(c^\star)
    =
    \left\{
        s_h(c^\star,\cdot)
    \right\}_{h=1}^{M},
    \label{eq:historical_hypotheses}
\end{equation}
where each hypothesis $h$ corresponds to a distinct way of transferring
historical experience to the target context.
For a candidate $z\in\mathcal{Z}_{\mathrm{cand}}$,
the scalar score $s_h(c^\star,z)$ measures how strongly hypothesis $h$
favors evaluating $z$ in context $c^\star$. Different hypotheses may assign different priorities to the same candidate. EVIS transfers historical \emph{ordering information} rather than treating historical predictions as target-environment decisions.

In our implementation, one source of such hypotheses is a collection
of context-specific Gaussian process classifiers
$s_i(z)\in[0,1]$, together with a pooled predictor
$s_{\mathrm{pool}}(z)$ fitted to all historical observations.
To reuse context-specific structure while accounting for variation
between environments, we additionally construct a retrieval-weighted
predictor.
Let $\phi(c)$ denote the descriptor used to compare contexts and let
$d_\phi$ denote the corresponding descriptor-space distance.
We define
\begin{equation}
    \begin{aligned}
        w_i(c^\star)
        &=
        \frac{
            \exp[-d_\phi(c^\star,c_i)/T]
        }{
            \sum_{j=1}^{N}
            \exp[-d_\phi(c^\star,c_j)/T]
        },\\
        s_{\mathrm{ret}}(c^\star,z)
        &=
        \sum_{i=1}^{N}
        w_i(c^\star)s_i(z),
    \end{aligned}
    \label{eq:historical_retrieval}
\end{equation}
where $T>0$ controls retrieval concentration.
The pooled and retrieval-based predictors provide complementary
hypotheses: the former emphasizes structure shared across environments,
whereas the latter places greater weight on historically similar
contexts.
Neither is assumed to transfer reliably to every target environment.

For multi-stage execution, historical trajectories provide richer
supervision than the final task outcome alone.
In addition to full-sequence success, intermediate events reveal
whether an initiation configuration supports the current skill and,
conditional on reaching the handoff state, whether downstream execution
can continue successfully.
We therefore construct separate historical hypotheses for local
feasibility, conditional continuation, and full-sequence completion.
These quantities provide complementary proposal signals: a
configuration that is favorable for the upstream skill need not be
favorable for the composed objective.
The corresponding scores are estimated from historical executions
using normalized RBF kernel smoothing, with continuation scores computed
only from trajectories that reach the relevant intermediate event.
We retain these scores as separate proposal hypotheses rather than
combining them into a single estimate of target-context success.

\textbf{Ranking-based proposal construction.}
Because historical predictors can differ substantially in numerical
scale and calibration, EVIS compares their relative preferences using
normalized percentile ranks:
\begin{equation}
    r_h(z)
    =
    \operatorname{PercentileRank}_{\mathcal{Z}_{\mathrm{cand}}}
    \bigl(s_h(c^\star,z)\bigr),
    \label{eq:hypothesis_rank}
\end{equation}
where larger values indicate more favorable configurations.
Any strictly increasing transformation of an individual score leaves
$r_h$ unchanged.
The proposal mechanism is therefore insensitive to the absolute scale
of a historical predictor and depends only on the ordering it induces.

EVIS converts the hypothesis-specific rankings into an ordered
proposal set by retaining high-ranking, non-duplicate candidates from
multiple historical hypotheses.
This avoids committing the search to a single transferred predictor:
candidates favored by different historical views are all considered
early in the search.
When additional historical proposals are needed, the ranked lists are
interleaved so that no single hypothesis dominates the query order.

Historical experience therefore determines where EVIS searches first,
but does not by itself determine which candidate is ultimately returned.
The latter decision is based on behavioral evidence collected in the
target environment.

\subsection{Behavioral Validation and Budgeted Return}
\label{subsec:assessment_termination}

\textbf{Repeated behavioral qualification.}
When an evaluated initiation configuration $z$ produces a successful
target-environment rollout, EVIS treats it as a provisional candidate
rather than returning it immediately.
Because a single rollout provides only one stochastic observation of
$p_G(c^\star,z)$, additional target-side evidence may be required before
the candidate is accepted.
When online behavioral validation is required, EVIS triggers fresh
executions from the same initiation configuration.
The discovery rollout that triggered validation is excluded from the
qualification statistics.
After $k$ fresh qualification rollouts, define
\begin{equation}
    S_k
    =
    \sum_{\ell=1}^{k}Y_G^{(\ell)},
    \qquad
    F_k=k-S_k,
    \label{eq:qualification_counts}
\end{equation}
where $S_k$ and $F_k$ denote the observed numbers of successes and
failures, respectively.

Given at most $n_q$ qualification trials and a required success count
$m_q$, EVIS applies the finite-budget decision rule
\begin{equation}
    \operatorname{Assess}_k(z)
    =
    \begin{cases}
        \mathrm{accept},
            & S_k\geq m_q,\\
        \mathrm{reject},
            & F_k>n_q-m_q,\\
        \mathrm{continue},
            & \text{otherwise, if budget remains}.
    \end{cases}
    \label{eq:qualification}
\end{equation}
The candidate is accepted as soon as $m_q$ successes have been
observed.
Conversely, it is rejected once the number of failures makes reaching
$m_q$ successes impossible within the remaining qualification trials.
These early stopping boundaries avoid executions that can no longer
change the final $m_q$-of-$n_q$ decision.

The qualification rule serves as a finite-budget behavioral screening
mechanism, using direct target-environment evidence to determine whether
a provisional candidate should be accepted.

\textbf{Interaction budget.}
Discovery and online qualification consume the same target-interaction
budget:
\begin{equation}
    K_{\mathrm{used}}
    =
    K_{\mathrm{search}}
    +
    K_{\mathrm{qual}}
    \leq
    K_{\max}.
    \label{eq:search_budget}
\end{equation}
If a provisional candidate is accepted, EVIS returns it immediately.
If it is rejected, the remaining interaction budget is passed to the
recovery search described in
Sec.~\ref{subsec:behavior_guided_search}.
If the budget is exhausted before a candidate receives the behavioral
support required by the active validation regime, EVIS abstains with
$\hat z=\bot$.

\subsection{Behavior-Guided Recovery Search}
\label{subsec:behavior_guided_search}

Historical proposals determine the initial search order, but imperfect
cross-context transfer may cause those proposals to fail or to be
rejected by behavioral validation.
EVIS therefore uses the target-side observations accumulated during
discovery to redirect the remaining search.

After $t$ discovery queries, let
\begin{equation}
    \mathcal{D}_t
    =
    \{(z_k,y_k)\}_{k=1}^{t},
    \qquad
    y_k=G(\xi_k),
    \label{eq:target_history}
\end{equation}
denote the target-side discovery history.
Let $\mathcal{V}_t$ denote the set of already queried configurations,
\begin{equation}
    \mathcal{U}_t
    =
    \mathcal{Z}_{\mathrm{cand}}
    \setminus
    \mathcal{V}_t
\end{equation}
the remaining candidates, and
$\mathcal{S}_t=\{z_k:y_k=1\}$ the configurations that have produced
successful discovery executions.
Recovery modifies where the search proceeds, while leaving the frozen
execution policies unchanged.

When historical proposals are insufficient, EVIS uses
$\mathcal{D}_t$ to redirect subsequent queries.

\textbf{Hypothesis disagreement.}
Different historical hypotheses may rank a candidate very differently,
which provides a useful signal for where transferred knowledge is
uncertain.
Let $\mathcal{H}_t$ denote the active scoring hypotheses and
$r_{h,t}$ their current candidate ranks.
EVIS defines
\begin{equation}
    a_{\mathrm{dis},t}(z)
    =
    \operatorname{Var}_{h\in\mathcal{H}_t}
    [r_{h,t}(z)]
    \left(
        \epsilon+
        \operatorname{Mean}_{h\in\mathcal{H}_t}
        [r_{h,t}(z)]
    \right),
    \label{eq:disagreement}
\end{equation}
where $\epsilon>0$ prevents candidates with moderate mean rank from
being completely suppressed.
The variance term favors candidates on which historical hypotheses
disagree, whereas the mean-rank term retains preference for candidates
that remain plausible under at least part of the historical evidence.
This quantity is used only as an exploration heuristic and is not
interpreted as calibrated uncertainty.

\textbf{Coverage and local exploration.}
When historical guidance does not produce sufficient evidence,
EVIS can broaden the search geometrically.
A coverage query selects the untested configuration farthest from the
existing discovery set:
\begin{equation}
    z_{\mathrm{cov}}
    =
    \arg\max_{z\in\mathcal{U}_t}
    \min_{u\in\mathcal{V}_t}
    \|\widetilde z-\widetilde u\|_2,
    \label{eq:coverage}
\end{equation}
where $\widetilde z$ denotes the normalized configuration coordinates.
This operator increases spatial coverage without making assumptions
about the calibration of historical models.
If successful configurations have already been observed, EVIS can
instead select an untested neighbor of a configuration in
$\mathcal{S}_t$, exploiting the empirical tendency of nearby initiation
states to share execution behavior.

\textbf{Regularized target correction.}
As additional target observations become available, EVIS can also
correct the transferred ranking without replacing the historical model.
Let
\begin{equation}
    s_H(z)
    =
    s_{\mathrm{ret}}(c^\star,z)
\end{equation}
denote the retrieval-based historical score.
We estimate a residual function
\begin{equation}
    \delta_t
    =
    \arg\min_{\delta\in\mathcal{H}_\kappa}
    \sum_{k=1}^{t}
    \bigl[
        y_k-s_H(z_k)-\delta(z_k)
    \bigr]^2
    +
    \lambda
    \|\delta\|_{\mathcal{H}_\kappa}^{2},
    \label{eq:residual}
\end{equation}
where $\mathcal{H}_\kappa$ is the RKHS associated with an RBF kernel and
$\lambda>0$ regularizes deviation from the transferred prediction.
The resulting target-corrected score is
\begin{equation}
    s_t(z)
    =
    \operatorname{clip}_{[\epsilon_s,\,1-\epsilon_s]}
    \bigl(
        s_H(z)+\delta_t(z)
    \bigr).
    \label{eq:target_score}
\end{equation}
The corrected score can participate in subsequent candidate ranking,
but, like the historical scores, is not interpreted as a calibrated
estimate of target success probability.

\begin{algorithm}[t]
    \caption{Experience-Guided and Behavior-Validated Initiation Search (EVIS)}
    \label{alg:evis}
    \begin{algorithmic}[1]
        \Require
        Historical data $\mathcal{D}_H$,
        target context $c^\star$,
        objective $G$,
        candidate set $\mathcal{Z}_{\mathrm{cand}}$,
        budget $K_{\max}$,
        configuration $\theta$
        \Ensure Candidate $\hat z$ or abstention $\bot$

        \State Construct historical hypotheses and ranked proposal queue
        \State $\mathcal{D}\gets\emptyset$,
               $\mathcal{V}\gets\emptyset$,
               $K\gets0$,
               $\mathrm{recover}\gets\mathrm{false}$

        \While{$K<K_{\max}$ and
               $\mathcal{Z}_{\mathrm{cand}}\setminus\mathcal{V}
               \neq\emptyset$}

            \If{$\neg\mathrm{recover}$}
                \Statex \hspace{\algorithmicindent}
                        \textit{Stage 1: Experience-guided proposal}
                \State Select next historical proposal $z$
            \Else
                \Statex \hspace{\algorithmicindent}
                        \textit{Stage 3: Behavior-guided recovery}
                \State $z\gets\mathrm{Recover}(\mathcal{D},\mathcal{V},\theta)$
            \EndIf

            \State Execute from $(c^\star,z)$ and observe $y=G(\xi)$
            \State $\mathcal{D}\gets\mathcal{D}\cup\{(z,y)\}$,
                   $\mathcal{V}\gets\mathcal{V}\cup\{z\}$,
                   $K\gets K+1$

            \If{$y=1$}
                \Statex \hspace{\algorithmicindent}
                        \textit{Stage 2: Behavioral validation}
                \State $(a,k_q)\gets
                       \mathrm{Validate}(z,K_{\max}-K,\theta)$
                \State $K\gets K+k_q$

                \If{$a=\mathrm{accept}$}
                    \State \Return $z$
                \ElsIf{$a=\mathrm{reject}$}
                    \State $\mathrm{recover}\gets\mathrm{true}$
                \EndIf
            \EndIf

            \If{historical proposals are insufficient}
                \State $\mathrm{recover}\gets\mathrm{true}$
            \EndIf

        \EndWhile

        \State \Return $\bot$
    \end{algorithmic}
\end{algorithm}

These operators serve a common purpose: once direct reuse of historical
proposals is no longer sufficient, target-side behavior determines how
the search expands.
The recovery stage may revisit historical rankings, explore regions of
high hypothesis disagreement, improve geometric coverage, exploit
local success structure, or use the regularized target correction,
according to the fixed search configuration.

Algorithm~\ref{alg:evis} summarizes the complete procedure.
Historical experience first constructs the proposal ordering;
target executions determine whether proposed configurations are
supported; and unsuccessful or rejected proposals trigger
behavior-guided recovery until a candidate is accepted or the
interaction budget is exhausted.

\section{Experiments}
\label{sec:experiments}

We evaluate EVIS at two complementary levels.
First, we assess initiation search on unseen target contexts
for both single-skill and multi-stage manipulation, measuring whether
historical guidance reduces target-environment interaction while
maintaining reliable candidate discovery.
We then conduct controlled mechanism analyses to isolate the roles of
experience-guided proposal, behavioral validation, and recovery search.

\subsection{Experimental Setup}
\label{subsec:experimental_setup}

We conduct experiments in RoboCasa365~\cite{nasiriany2026robocasa365}.
The learned skills are instantiated using the pretrained
RLDX-1-FT-RC365 Vision-Language-Action (VLA) policy~\cite{rldx2026},
whose parameters remain frozen throughout evaluation.
A context is a fixed environment instance specified by its kitchen
layout, visual style, and initialization.

\textbf{Initiation space and tasks.}
We parameterize the robot base pose relative to the target fixture as
$z=(\Delta x,\Delta y,\Delta\psi)$, where
$\Delta x$ and $\Delta y$ denote the planar position offsets and
$\Delta\psi$ denotes the yaw offset relative to the fixture.
The search ranges are
$\Delta x\in[-0.25,0.25]$\,m,
$\Delta y\in[-0.20,0.20]$\,m, and
$\Delta\psi\in[-30^\circ,30^\circ]$. We consider \textit{OpenDrawer} as the single-skill task.
For multi-stage evaluation, \textit{OpenDrawer} is followed by either
\textit{DrawerUtensilSort} or \textit{ChooseMeasuringCup}.
The downstream skill begins from the physical state produced when the
drawer-accessibility condition is reached.
The environment is not reset between stages, and success requires
completion of the entire skill sequence.

\begin{figure}[t]
    \centering
    \includegraphics[width=8.6cm]{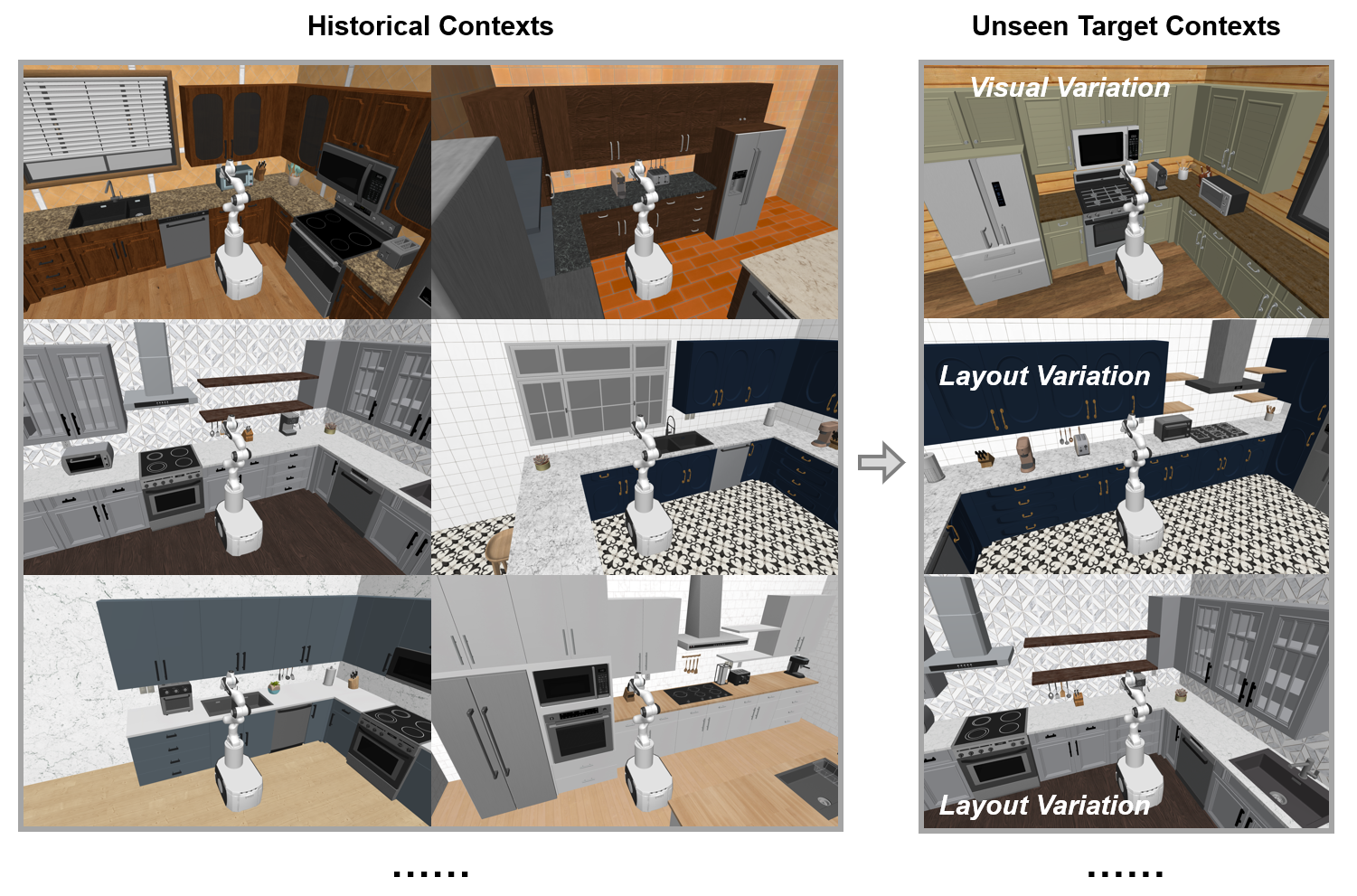}
    \caption{Representative historical contexts and unseen target contexts in
    RoboCasa365, illustrating variations in visual appearance and kitchen layout.}
    \label{fig:context_examples}
    \vspace{-0.3cm}
\end{figure}

\textbf{Historical and target contexts.}
We divide the environments into \emph{historical contexts} and
\emph{unseen target contexts}.
Historical contexts provide the execution experience used to construct
historical predictors, select proposal sources, and determine search
hyperparameters, while unseen target contexts are reserved for
evaluation.
Fig.~\ref{fig:context_examples} shows representative examples from the
two sets, highlighting variations in both visual appearance and kitchen
layout.

\textbf{Metrics.}
\emph{Mean Queries} is the average number of target rollouts used before
termination, including discovery and online qualification but excluding
independent confirmation.
\emph{Confirmed SR} is the macro-average confirmation success rate,
assigning zero to abstention.
\emph{Reliable Rate} is the fraction of target contexts whose returned
candidates satisfy the independent reliability criterion.
For selective-return analyses, \emph{False Accepts} counts returned but
unreliable candidates, while \emph{Reliable Precision} is the fraction
of returned candidates that are reliable.
\emph{Recovery Rate} is the fraction of contexts entering recovery for
which subsequent search returns a reliable candidate.

\subsection{Initiation Search on Unseen Target Contexts}
\label{subsec:unseen_target_search}

We first evaluate initiation search on unseen target contexts
for both single-skill and multi-stage manipulation.

\textbf{Single-skill discovery.}
We compare EVIS with a matched fixed-budget variant, a five-query
historical-reuse baseline, and Scratch-Sobol.
EVIS and Fixed B20 use the same query-generation and return rules,
while Fixed B20 always consumes the full 20-query budget.
History Reuse B5 returns after the five initial history-informed probes.
Scratch-Sobol B20 ignores historical experience and queries candidates
according to a Sobol low-discrepancy sequence, which provides
space-filling coverage of the search domain~\cite{sobol1967distribution}.

\begin{figure}[!t]
    \centering
    \includegraphics[width=\columnwidth]
    {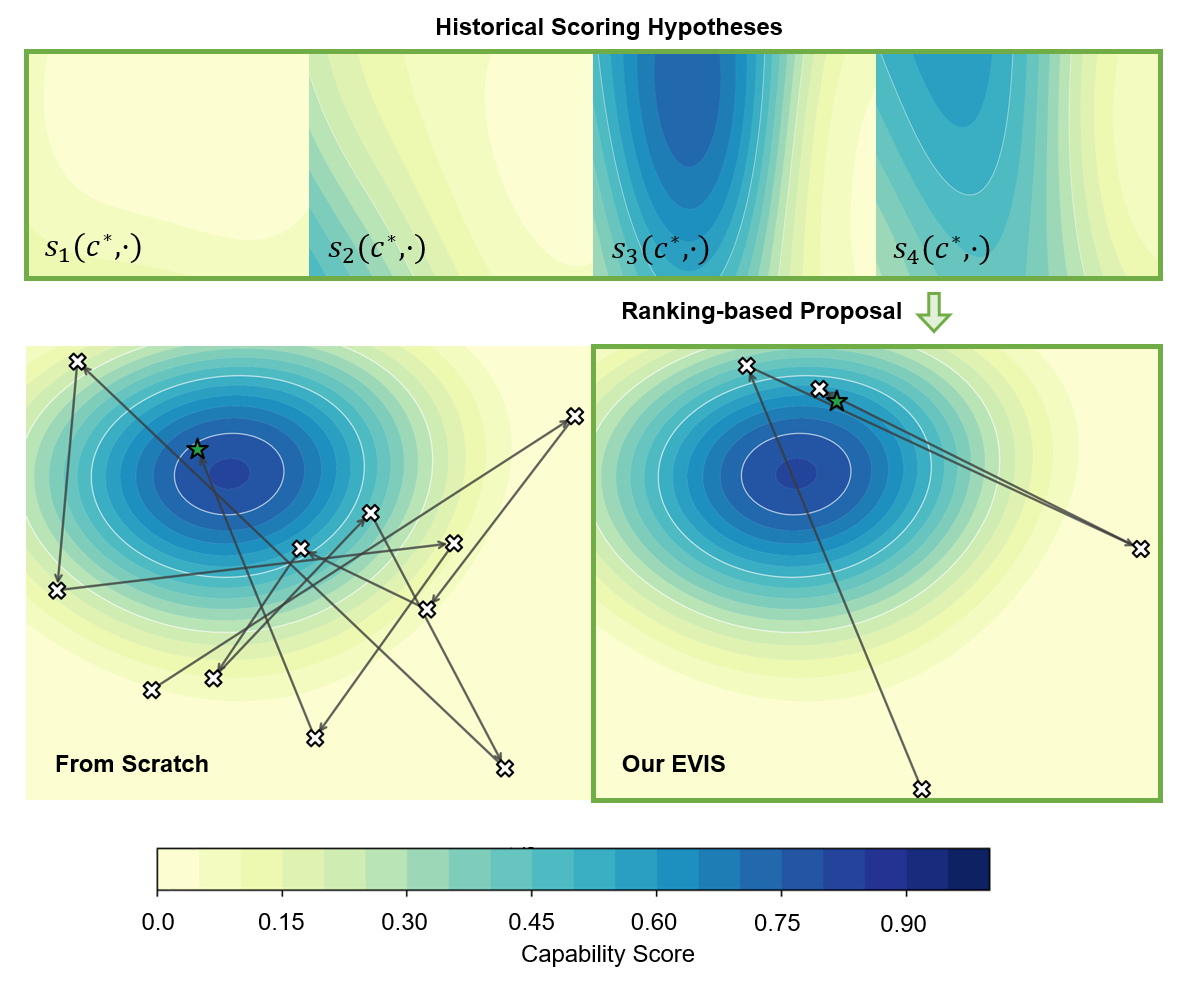}
    \caption{Representative single-skill initiation search on an unseen
    target context. Historical guidance prioritizes promising
    configurations, allowing EVIS to observe a successful candidate
    earlier than Scratch-Sobol.}
    \label{fig:single_stage_spatial_case}
    \vspace{-0.2cm}
\end{figure}

\begin{table}[t]
\centering
\caption{Single-skill initiation search.}
\label{tab:single_skill}
\resizebox{\columnwidth}{!}{
\begin{tabular}{@{}lcccc@{}}
\toprule
Metric
& EVIS
& EVIS-Fixed B20
& History Reuse B5
& Scratch-Sobol B20 \\
\midrule
Mean Queries $\downarrow$
& 8.125
& 20.000
& \textbf{5.000}
& 20.000 \\

Confirmed SR $\uparrow$
& \textbf{76.7\%}
& \textbf{76.7\%}
& 70.8\%
& 72.5\% \\
\bottomrule
\end{tabular}
}
\vspace{-0.1cm}
\end{table}

EVIS returns the same candidate as EVIS-Fixed B20 while using only $8.125$ queries on average, reducing target
interaction by $59.4\%$ with identical confirmation outcomes.
This shows that behavior-dependent stopping removes unnecessary
interaction without changing the fixed-budget output.
Compared with History Reuse B5, EVIS uses additional target interaction
but improves Confirmed SR from $70.8\%$ to $76.7\%$, indicating the
benefit of continued search beyond direct historical reuse.

Fig.~\ref{fig:single_stage_spatial_case} provides a representative
example.
Historical guidance prioritizes a promising region, and EVIS observes
its first successful initiation configuration at query 5, whereas
Scratch-Sobol requires 12 queries in this context.

\textbf{Multi-stage initiation search.}
We next evaluate initiation search when candidate quality is determined
by completion of the entire skill sequence.
To isolate the value of historical guidance under matched interaction
budgets, we compare EVIS with Scratch-Sobol using the same budget,
qualification rule, candidate pool, and independent confirmation
protocol.
Reliable Rate is the primary measure of successful candidate discovery,
with Confirmed SR reported as a complementary measure of execution
quality.

\begin{table*}[t]
\centering
\caption{Multi-stage initiation search.}
\label{tab:multi_stage}
\resizebox{\textwidth}{!}{
\begin{tabular}{@{}c cccc cccc@{}}
\toprule
\multirow{3}{*}{Budget $B$}
& \multicolumn{4}{c}{
$\rightarrow$\textit{DrawerUtensilSort}}
& \multicolumn{4}{c}{
$\rightarrow$\textit{ChooseMeasuringCup}} \\
\cmidrule(lr){2-5}
\cmidrule(lr){6-9}
& \multicolumn{2}{c}{EVIS}
& \multicolumn{2}{c}{Scratch-Sobol}
& \multicolumn{2}{c}{EVIS}
& \multicolumn{2}{c}{Scratch-Sobol} \\
\cmidrule(lr){2-3}
\cmidrule(lr){4-5}
\cmidrule(lr){6-7}
\cmidrule(lr){8-9}
& Reliable Rate & Confirmed SR
& Reliable Rate & Confirmed SR
& Reliable Rate & Confirmed SR
& Reliable Rate & Confirmed SR \\
\midrule

6
& \textbf{25.0\%} & \textbf{12.1\%}
& 0.0\% & 0.0\%
& \textbf{33.3\%} & \textbf{14.6\%}
& 8.3\% & 3.8\% \\

10
& \textbf{41.7\%} & \textbf{20.4\%}
& 8.3\% & 5.4\%
& \textbf{58.3\%} & \textbf{29.2\%}
& 16.7\% & 7.1\% \\

15
& \textbf{41.7\%} & \textbf{20.4\%}
& 33.3\% & 20.0\%
& \textbf{66.7\%} & \textbf{33.3\%}
& 41.7\% & 17.9\% \\

20
& \textbf{66.7\%} & \textbf{32.9\%}
& 58.3\% & 30.8\%
& \textbf{66.7\%} & \textbf{33.3\%}
& 58.3\% & 24.6\% \\

\bottomrule
\end{tabular}
}
\vspace{-0.2cm}
\end{table*}

The benefit of EVIS is most pronounced under small interaction budgets.
At $B=10$, EVIS achieves Reliable Rates of $41.7\%$ and $58.3\%$ on
the two skill sequences, compared with $8.3\%$ and $16.7\%$ for
Scratch-Sobol.
The gap narrows as additional target interaction becomes available.
These results suggest that the primary benefit of historical guidance
lies in improving early candidate discovery under limited budgets, with
diminishing gains as target-side exploration accumulates.

Importantly, successful execution of the upstream skill does not imply
successful completion of the composed task.
For \textit{OpenDrawer}$\rightarrow$\textit{ChooseMeasuringCup},
$37.5\%$ of evaluated executions that successfully completed
\textit{OpenDrawer} still failed during downstream execution.
This result motivates evaluating initiation configurations with respect
to the full skill sequence rather than upstream skill success alone.

\subsection{Mechanism Analysis: Propose, Validate, and Recover}
\label{subsec:mechanism_analysis}

We next isolate the contributions of experience-guided proposal,
behavioral validation, and recovery search.

\textbf{Experience-guided proposal.}
We first examine whether historical experience provides useful
cross-context ordering before target-side validation is applied.
For single-skill drawer opening, we evaluate how well historical
scores rank promising initiation configurations across unseen target
contexts.
For multi-stage initiation search, we then compare the same
target-adaptive optimizer with and without historical initialization on
unseen \textit{OpenDrawer}$\rightarrow$\textit{ChooseMeasuringCup}
contexts.

\begin{figure}[!t]
    \centering
    \includegraphics[width=\columnwidth]
    {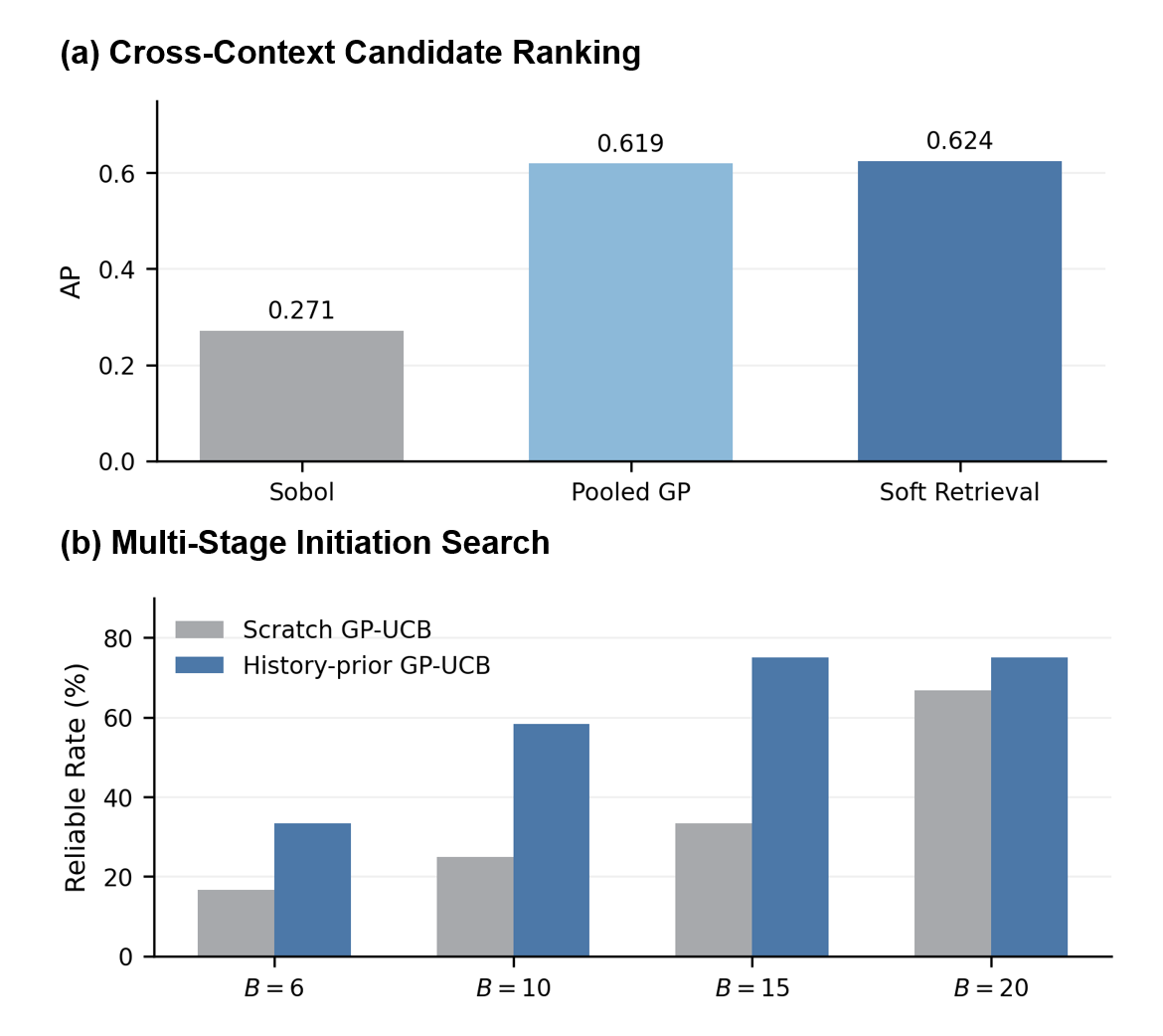}
    \caption{Effect of experience-guided proposals.
    (a) Cross-context candidate ranking for single-skill drawer opening.
    (b) Multi-stage initiation search with and without historical
    initialization under increasing target-interaction budgets.}
    \label{fig:history_effect}
    \vspace{-0.3cm}
\end{figure}

As shown in Fig.~\ref{fig:history_effect}(a), historical experience
provides transferable ranking structure.
Both Pooled GP and Soft Retrieval achieve an Average Precision (AP)
above $0.61$, compared with $0.271$ for the history-agnostic Sobol
ordering.

This ranking advantage also improves downstream search efficiency
(Fig.~\ref{fig:history_effect}(b)).
Under the same target-adaptive GP-UCB optimizer, historical
initialization increases reliable candidate discovery from $25.0\%$
to $58.3\%$ at $B=10$ and from $33.3\%$ to $75.0\%$ at $B=15$.
The gap narrows at $B=20$ as additional target-side exploration
accumulates.
These results support the role of historical experience as a proposal
mechanism: it prioritizes promising configurations early without
determining the final returned candidate.

\textbf{Behavioral validation.}
Historical proposals can prioritize promising initiation configurations,
but their reliability in a new target context cannot be inferred from
history alone.
To isolate the role of target-side validation, we fix the same historical
candidate within each task and compare direct reuse with the 5-of-6
qualification rule, without additional search.

\begin{table}[t]
\centering
\caption{Effect of behavioral validation.}
\label{tab:validation_ablation}
\resizebox{\columnwidth}{!}{
\begin{tabular}{@{}llcc@{}}
\toprule
Task
& Method
& False Accept $\downarrow$
& Rel. Precision $\uparrow$ \\
\midrule

\multirow{2}{*}{ChooseMeasuringCup}
& History-only
& 7
& 41.7\% \\
& + Qualification
& \textbf{0}
& \textbf{100\%} \\

\midrule

\multirow{2}{*}{DrawerUtensilSort}
& History-only
& 11
& 8.3\% \\
& + Qualification
& \textbf{0}
& \textbf{100\%} \\

\bottomrule
\end{tabular}
}
\vspace{-0.2cm}
\end{table}

Table~\ref{tab:validation_ablation} shows that direct historical reuse
frequently returns candidates that are unreliable in the target context.
Adding target-side qualification eliminates all false accepts in both
tasks, increasing reliable precision from $41.7\%$ and $8.3\%$ to
$100\%$.
This gain in precision comes at the cost of reduced return coverage,
because rejected candidates are not replaced in this ablation.
The result therefore highlights the distinct roles of proposal and
validation: history determines where to try, while target-side behavior
determines whether a transferred candidate should be trusted.

\textbf{Recovery after rejected reuse.}
Behavioral validation may leave target contexts unresolved when the
historical candidate is rejected, so we test whether continued search
can recover reliable configurations.
All methods use the same fixed historical proposal and 5-of-6
qualification stage: Qualification Only stops after rejection, while
Sobol Fallback and EVIS Fallback resume with history-agnostic Sobol
search and history-guided EVIS search, respectively.

\begin{table}[t]
\centering
\caption{Recovery after rejected historical reuse.}
\label{tab:recovery_cascade}
\resizebox{\columnwidth}{!}{
\begin{tabular}{@{}lccc@{}}
\toprule
Method
& Recovery Rate $\uparrow$
& Final Reliable Rate $\uparrow$
& Mean Queries $\downarrow$ \\
\midrule

Qualification Only
& 0.0\%
& 8.3\%
& 2.33 \\

Sobol Fallback
& 42.9\%
& 33.3\%
& 13.83 \\

EVIS Fallback
& \textbf{57.1\%}
& \textbf{41.7\%}
& 13.00 \\

\bottomrule
\end{tabular}
}
\vspace{-0.2cm}
\end{table}

Table~\ref{tab:recovery_cascade} shows that fallback search restores
coverage after rejected reuse, increasing the final reliable rate from
$8.3\%$ to $33.3\%$ with Sobol fallback and $41.7\%$ with EVIS-guided
fallback.
EVIS-guided recovery also uses slightly fewer target queries.
Together, these results support the propose--validate--recover design:
history prioritizes where to search, target-side behavior determines
what to trust, and continued search restores coverage after failed
transfer.
\section{Conclusion}
\label{sec:conclusion}

We presented EVIS, an Experience-Guided and Behavior-Validated
Initiation Search framework for discovering reliable initiation
configurations of frozen learned skills under limited target-environment
interaction.
EVIS treats historical execution experience as guidance for where to
search, while target-side behavioral evidence determines whether a
candidate should be trusted and whether additional search is needed.
Our experiments show that this combination improves reliable candidate
discovery under limited interaction budgets and reduces the number of
target-environment rollouts required for deployment.
These results demonstrate that historical experience can be reused
effectively without assuming perfect transfer to a new environment.


{
    \bibliographystyle{IEEEtran}
    \bibliography{IEEEabrv, bib/bibliography}

\begin{thebibliography}{10}
\providecommand{\url}[1]{#1}
\csname url@samestyle\endcsname
\providecommand{\newblock}{\relax}
\providecommand{\bibinfo}[2]{#2}
\providecommand{\BIBentrySTDinterwordspacing}{\spaceskip=0pt\relax}
\providecommand{\BIBentryALTinterwordstretchfactor}{4}
\providecommand{\BIBentryALTinterwordspacing}{\spaceskip=\fontdimen2\font plus
\BIBentryALTinterwordstretchfactor\fontdimen3\font minus \fontdimen4\font\relax}
\providecommand{\BIBforeignlanguage}[2]{{%
\expandafter\ifx\csname l@#1\endcsname\relax
\typeout{** WARNING: IEEEtran.bst: No hyphenation pattern has been}%
\typeout{** loaded for the language `#1'. Using the pattern for}%
\typeout{** the default language instead.}%
\else
\language=\csname l@#1\endcsname
\fi
#2}}
\providecommand{\BIBdecl}{\relax}
\BIBdecl

\bibitem{kim2024openvla}
M.~J. Kim, K.~Pertsch, S.~Karamcheti, T.~Xiao, A.~Balakrishna, S.~Nair, R.~Rafailov, E.~Foster, G.~Lam, P.~Sanketi \emph{et~al.}, ``Openvla: An open-source vision-language-action model,'' \emph{arXiv preprint arXiv:2406.09246}, 2024.

\bibitem{huang2026roboharness}
J.~Huang, Y.~Hu, Z.~Li, R.~Qi, Y.~Xiao, Z.~Zhang, M.~Coates, T.~Cao, and Y.~Zhang, ``Roboharness: Memory-driven orchestration of heterogeneous robot policies for long-horizon planning,'' \emph{arXiv preprint arXiv:2607.18060}, 2026.

\bibitem{zhang2026harness}
Y.~Zhang, H.~Zhang, F.~Gao, X.~Li, Z.~Liu, C.~Zhu, J.~Qiu, Y.~Yan, J.~Liu, W.~Tang \emph{et~al.}, ``Harness vla: Steering frozen vlas into reliable manipulation primitives via memory-guided agents,'' \emph{arXiv preprint arXiv:2607.08448}, 2026.

\bibitem{huang2022inner}
W.~Huang, F.~Xia, T.~Xiao, H.~Chan, J.~Liang, P.~Florence, A.~Zeng, J.~Tompson, I.~Mordatch, Y.~Chebotar \emph{et~al.}, ``Inner monologue: Embodied reasoning through planning with language models,'' \emph{arXiv preprint arXiv:2207.05608}, 2022.

\bibitem{garrett2021integrated}
C.~R. Garrett, R.~Chitnis, R.~Holladay, B.~Kim, T.~Silver, L.~P. Kaelbling, and T.~Lozano-P{\'e}rez, ``Integrated task and motion planning,'' \emph{Annual review of control, robotics, and autonomous systems}, vol.~4, no.~1, pp. 265--293, 2021.

\bibitem{garrett2020pddlstream}
C.~R. Garrett, T.~Lozano-P{\'e}rez, and L.~P. Kaelbling, ``Pddlstream: Integrating symbolic planners and blackbox samplers via optimistic adaptive planning,'' in \emph{Proceedings of the international conference on automated planning and scheduling}, vol.~30, 2020, pp. 440--448.

\bibitem{liang2022learning}
J.~Liang, X.~Cheng, and O.~Kroemer, ``Learning preconditions of hybrid force-velocity controllers for contact-rich manipulation,'' \emph{arXiv preprint arXiv:2206.12728}, 2022.

\bibitem{sharma2020relational}
M.~Sharma and O.~Kroemer, ``Relational learning for skill preconditions,'' \emph{arXiv preprint arXiv:2012.01693}, 2020.

\bibitem{nasiriany2026robocasa365}
S.~Nasiriany, S.~Nasiriany, A.~Maddukuri, and Y.~Zhu, ``Robocasa365: A large-scale simulation framework for training and benchmarking generalist robots,'' in \emph{International Conference on Learning Representations}, vol. 2026, 2026, pp. 98\,643--98\,667.

\bibitem{rldx2026}
D.~Kim, H.~Jang, M.~Koo, S.~Jang, T.~Kim \emph{et~al.}, ``Rldx-1 technical report,'' \emph{arXiv preprint arXiv:2605.03269}, 2026.

\bibitem{sutton1999between}
R.~S. Sutton, D.~Precup, and S.~Singh, ``Between mdps and semi-mdps: A framework for temporal abstraction in reinforcement learning,'' \emph{Artificial intelligence}, vol. 112, no. 1-2, pp. 181--211, 1999.

\bibitem{konidaris2009skill}
G.~Konidaris and A.~Barto, ``Skill discovery in continuous reinforcement learning domains using skill chaining,'' \emph{Advances in neural information processing systems}, vol.~22, 2009.

\bibitem{ahn2022can}
M.~Ahn, A.~Brohan, N.~Brown, Y.~Chebotar, O.~Cortes, B.~David, C.~Finn, C.~Fu, K.~Gopalakrishnan, K.~Hausman \emph{et~al.}, ``Do as i can, not as i say: Grounding language in robotic affordances,'' \emph{arXiv preprint arXiv:2204.01691}, 2022.

\bibitem{liang2023code}
J.~Liang, W.~Huang, F.~Xia, P.~Xu, K.~Hausman, B.~Ichter, P.~Florence, and A.~Zeng, ``Code as policies: Language model programs for embodied control,'' in \emph{2023 IEEE International conference on robotics and automation (ICRA)}.\hskip 1em plus 0.5em minus 0.4em\relax IEEE, 2023, pp. 9493--9500.

\bibitem{cully2015robots}
A.~Cully, J.~Clune, D.~Tarapore, and J.-B. Mouret, ``Robots that can adapt like animals,'' \emph{Nature}, vol. 521, no. 7553, pp. 503--507, 2015.

\bibitem{pautrat2018bayesian}
R.~Pautrat, K.~Chatzilygeroudis, and J.-B. Mouret, ``Bayesian optimization with automatic prior selection for data-efficient direct policy search,'' in \emph{2018 IEEE International Conference on Robotics and Automation (ICRA)}.\hskip 1em plus 0.5em minus 0.4em\relax IEEE, 2018, pp. 7571--7578.

\bibitem{feurer2018scalable}
M.~Feurer, B.~Letham, and E.~Bakshy, ``Scalable meta-learning for bayesian optimization using ranking-weighted gaussian process ensembles,'' in \emph{AutoML workshop at ICML}, vol.~7, 2018, p.~5.

\bibitem{locatelli2016optimal}
A.~Locatelli, M.~Gutzeit, and A.~Carpentier, ``An optimal algorithm for the thresholding bandit problem,'' in \emph{International Conference on Machine Learning}.\hskip 1em plus 0.5em minus 0.4em\relax PMLR, 2016, pp. 1690--1698.

\bibitem{kano2019good}
H.~Kano, J.~Honda, K.~Sakamaki, K.~Matsuura, A.~Nakamura, and M.~Sugiyama, ``Good arm identification via bandit feedback,'' \emph{Machine Learning}, vol. 108, no.~5, pp. 721--745, 2019.

\bibitem{jourdan2026anytime}
M.~Jourdan, A.~Delahaye-Duriez, and C.~R{\'e}da, ``An anytime algorithm for good arm identification,'' \emph{Journal of Machine Learning Research}, vol.~27, no.~19, pp. 1--90, 2026.

\bibitem{sobol1967distribution}
I.~M. Sobol, ``Distribution of points in a cube and approximate evaluation of integrals,'' \emph{USSR Computational mathematics and mathematical physics}, vol.~7, pp. 86--112, 1967.

\end{thebibliography}
}

\end{document}